\pdfoutput=1

\documentclass[11pt]{article}

\usepackage[final]{acl}
\usepackage{zref}
\usepackage{times}
\usepackage{latexsym}

\usepackage[T1]{fontenc}

\usepackage[utf8]{inputenc}

\usepackage{microtype}

\usepackage{inconsolata}

\usepackage{graphicx}
\usepackage{multicol}
\usepackage{float}
\title{On Fact and Frequency: LLM Responses to Misinformation Expressed with Uncertainty}

\author{
  \textbf{Yana van de Sande \textsuperscript{1,2}},
  \textbf{Gunes Acar\textsuperscript{2,3}},
  \textbf{Thabo van Woudenberg\textsuperscript{1}},
  \textbf{Martha Larson\textsuperscript{1,3}},
\\
  \textsuperscript{1}Centre for Language Studies
  \textsuperscript{2}iHub
  \textsuperscript{3}Institute for Computing and Information Sciences
\\
Radboud University
\\
\\
  \small{
    \textbf{Correspondence:} \href{mailto:yana.vandesande@ru.nl}{yana.vandesande@ru.nl}
  }
}

\begin{document}
\maketitle
\begin{abstract}
We study LLM judgments of misinformation expressed with uncertainty.
Our experiments study the response of three widely used LLMs (GPT-4o, LLaMA3, DeepSeek-v2) to misinformation propositions that have been verified false and then are transformed into uncertain statements according to an uncertainty typology.
Our results show that after transformation, LLMs change their factchecking classification from false to not-false in 25\% of the cases.
Analysis reveals that the change cannot be explained by predictors to which humans are expected to be sensitive, i.e., modality, linguistic cues, or argumentation strategy.
The exception is doxastic transformations, which use linguistic cue phrases such as “It is believed…”.
To gain further insight, we prompt the LLM to make another judgment about the transformed misinformation statements that is not related to truth value.
Specifically, we study LLM estimates of the frequency with which people make the uncertain statement.
We find a small but significant correlation between judgment of fact and estimation of frequency.

\end{abstract}

\section{Introduction}
\begin{figure}[t]
    \centering
    \includegraphics[width=0.9\linewidth]{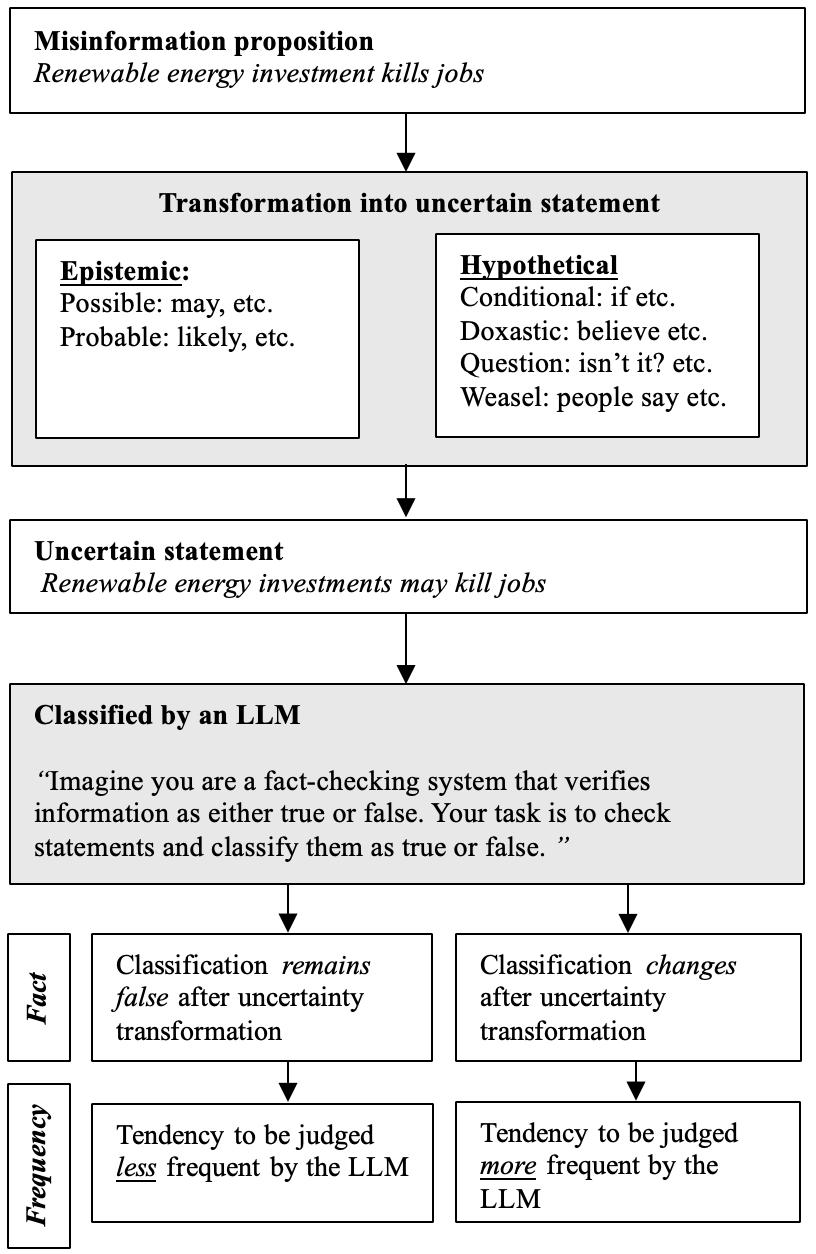}
    \caption{Overview of our experiments and findings on LLM responses to misinformation that is expressed uncertainly.}
    \label{fig:summary}
\end{figure}
Large language models (LLMs) 
offer promising support for fact checking information, but come with technical and societal challenges~\cite{quelle_perils_2024, augenstein_factuality_2024}. 
Most research focuses on LLM classification for statements that are clearly propositions, meaning that they can be assigned a truth value, either true or false. 
However, misinformation experts caution against more deceptive forms of misinformation (including disinformation) that fall into the realm of uncertainty, where truth values cannot be clearly assigned ~\cite{van2024broader}. 
Uncertainty is in the broadest sense defined as a lack of information and impacts the attribution a truth value, which has implications for the judgment of the reliability of a statement~\cite{szarvas_cross-genre_2012}.

Recent work has explored the mechanisms of uncertainty within LLMs. 
The focus has remained relatively narrow.
Work has studied how LLMs represent uncertainty internally e.g.,~\cite{belem_perceptions_2024},
or looked at effect of uncertainty on language generation~\cite{zhou_navigating_2023}. 
Such work established that uncertainty affects the ability of LLMs to identify facts.
Our work is positioned in the intersection of uncertainty research on LLMs and work that looks at LLMs as aid for fact checking. It explores how LLMs respond to misinformation that is expressed with uncertainty.
The larger motivation of our research is the concern that uncertain statements might either escape fact checkers or be unfairly flagged, thus restricting free speech.

A summary of our experiments and findings is provided in Figure~\ref{fig:summary}. 
We start by creating a dataset of misinformation expressed with uncertainty
that is based on verified-false propositions in the  widely used ClimateFever dataset~\cite{diggelmann_climate-fever_2021}. 
We then prompt three popular LLMs (GPT-4o, LLaMA3, DeepSeek-v2) to fact-check the uncertain statements and measure how uncertainty transformation changes their classification. 
We find that transformations cause a significant change in classification of misinformation from false to not-false labels in 25\% of the cases. 
We further explore the response of LLMs to uncertainly-expressed information by investigating LLMs estimates of the frequency at which people make statements of this specific type in this specific context.
The aim of studying frequency is to gain understanding of LLMs response to uncertainty with respect to a type of judgment beyond a fact-checking judgment.
We find a tendency for the LLM to judge a statement more frequent if its classification is also changed after transformation. 

Our work makes the following contributions:
\begin{itemize}
\vspace{-2mm}
\setlength\itemsep{-0.2em}
    \item A typology of uncertainty expressions based on different corpora annotations to evaluate how linguistic features impact LLM classification. 
    \item A dataset of verified-false misinformation propositions transformed into uncertain statements using these categories.
    \item An analysis of how the classification of a LLM changes when misinformation is expressed with uncertainty.
    \item A prompt to elicit frequency estimates from LLMs on uncertain information, and an exploratory analysis of these estimates. 
\end{itemize}
Uncertainty typology, data set, and the code for this paper can be found at \href{https://github.com/yanavdsande/on-fact-and-frequency}{yanavdsande/on-fact-and-frequency}

\section{Background on uncertainty}
\label{sec:background}


The study of uncertainty has given rise to a broad body of linguistic literature. 
Two concepts that appear widely in these studies are modalities and linguistic constructions through cue phrases.
Modalities are ways that humans associate meaning with information. 
\citet{palmer2001mood} proposed a modality framework as a cross-language grammatical category. The modality famework is connected to uncertainty through epistemic and hypothetical modalities \cite{szarvas_cross-genre_2012, cruz2019negation}. 
The difference between the two modalities is how uncertainty emerges in the absence or presence of information.
Epistemic uncertainty occurs when there is not enough information to determine if a statement is true or false. 
In hypothetical uncertainty, a statement can be both true and false at the same time, so knowing its truth value does not resolve the uncertainty.

Linguistic subcategories are ways in which humans differentiate types of uncertainty within modalities. 
Different authors have identified linguistic subcategories and the cue phrases that signal them.
\citet{szarvas_cross-genre_2012} constructed a domain- and genre independent taxonomy of semantic uncertainty, mapping linguistic cues on the philosophical modalities. 
They distinguish two linguistic subcategories to the epistemic modality that are also observed by~\citet{sauri2009factbank}: possible ("may") and probable ("could").
The difference between the two being certainty attributed to the likelihood of a statement (probability giving a higher likelihood of the statement reflecting the actual world than possibility).  
Within the hypothetical modality,~\citet{szarvas_cross-genre_2012} state linguistic cue phrases can take the form of conditionals ("if - then" etc.), doxastic constructions ("I believe" etc.), and investigative indicators ("examine" etc.).  
In a continuation of the work by \citet{szarvas_cross-genre_2012}, \citet{wei_empirical_2013} used the categorizations of uncertainty to research uncertainty expression in microblogs on social media. 
They found that on social media uncertainty is not often expressed by linguistic markers from the investigative category, and updated the framework with the inclusion of questions ("isn't it") and external references. 
External references is the vague attribution of a source, where a statement is dressed with authority, yet has no substantial basis\cite{wei_empirical_2013}. The notion of external references as a cause of uncertainty has also been mentioned within the context of weasels \cite{vincze2013weasels}. 


\section{Experimental Setup}

\subsection{Dataset}
\begin{figure}
    \centering
    \includegraphics[width=0.9\linewidth]{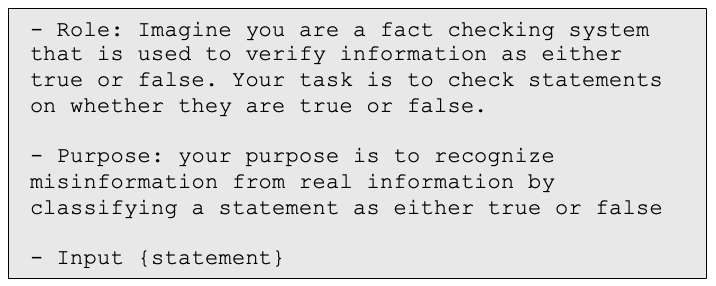}
    \caption{Prompt used for false / not-false classification}
    \label{figprompt}
\end{figure}

We carry out our experiments on a dataset of misinformation statements that were also classified as misinformation by the LLMs that we study.
In order to base our statements on verified misinformation, we build our data set on verified-false (i.e., misinformation) propositions in the ClimateFever data set~\cite{diggelmann_climate-fever_2021}.
We chose the verified-false propositions that are understandable in isolation and not part of a larger linguistic context, leaving us with 211 propositions (of the original 254). 
We zero-shot prompt (Figure \ref{figprompt}) the LLMs to classify the 211 propositions on being false or true.
Details of classification are described Section~\ref{subsec:models} and Sections~\ref{subsec:fvsnf} \& ~\ref{subsec:frequency_elicit}. Only propositions correctly classified as false were selected.
The result was 181 propositions for GPT-4o, 136 propositions for LLaMA3 and 110 propositions for DeepSeek-v2.). 
We then transform the misinformation propositions into uncertain statements by applying uncertainty transformations, resulting in our final data set of misinformation expressed with uncertainty.


In order to construct a dataset of uncertainty transformations, we investigate uncertainty labels in datasets from the CONLL-2010 shared task~\cite{farkas2010conll} on detecting uncertain information, (FactBank \cite{sauri2009factbank}, Wikiweasel \cite{vincze2013weasels}, BioScope \cite{vincze2008bioscope}).
We extract cues corresponding to the epistemic and hypothethical modalities as to the linguistic subcategories mentioned in Section~\ref{sec:background}.
In our work we include the external category as observed by \citet{wei_empirical_2013}, even though it was not present in the CONLL-2010 shared task. 
We follow the work by \citet{vincze2013weasels} and \citet{zhou_navigating_2023} to construct linguistic cues for this category. 
We extract the top 10 most-used categorically unique lexical cues to be included for transformation. 
For more information on the transformation see appendix \ref{app-transformation}. 

\subsection{Models}
\label{subsec:models}

We test three popular models: GPT-4o-mini:8b, LLaMA3:8b, and DeepSeek-v2:16b. 
For the classification we use a temperature setting of 0.0, to make the output more deterministic \cite{mirza_global-liar_2024, hoes2023leveraging, zhou_navigating_2023}. 
We use a zero-shot prompting method on our dataset. 
We prompt the model 5 times for each data-point. For each query, the context given to the model only includes our prompt and the statement (Figure \ref{figprompt}), precluding that the models' classification is influenced by previous classifications.


\subsection{False vs. not-false classification} 
\label{subsec:fvsnf}
Propositions and transformed statements are classified similarly.
The prompt used (Figure \ref{figprompt}) was constructed based on the instructions by OpenAI and Meta on prompt design, and on work exploring LLMs in the context of fact checking \cite{mirza_global-liar_2024, hoes2023leveraging}. 
We refrain from hard forcing the model to classify through an output template. 
We construct a majority vote scheme based on the five outputs by the LLM per data point, where the majority classification is considered the classification that is used to calculate the accuracy.  
For the classification of uncertain statements, all predictions that are not false are considered different classification responses. 
We conduct a McNemars test to assess whether the LLM responds significantly different to uncertain statements than to facts. 

\subsection{Frequency elicitation}
\label{subsec:frequency_elicit}
\begin{figure}
    \centering
    \includegraphics[width=0.9\linewidth]{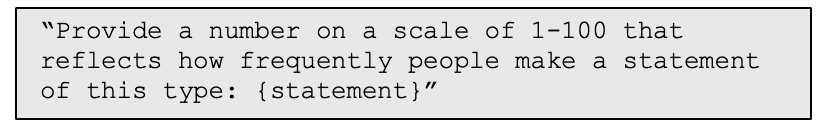}
    \caption{Prompt used to elicit frequency}
    \label{fig_prompt_freq}
\end{figure}

For frequency elicitation, we prompt the LLMs to express the frequency with which people make statements of a similar type on a scale from 0 to 100 using the prompt in Figure \ref{fig_prompt_freq}. 
In contrast to the prompt used for classification, this prompt was not drawn from the literature, since to our knowledge, we are the first to elicit this type of frequency information from an LLM.
For all statements, we got a numeric response between 1 and 100, of which none were 1 and none were 100. 
We prompt each statement five times and the five numerical values were averaged for use in the analysis.
We conduct a logistic regression to analyze whether LLM factchecking classification can be predicted by elicited frequency. 

\section{Results}
Table~\ref{first_experiment_confusion matrix} presents our classification results at a high level.
For all three models, ca. 75\% of the statements are still classified as false after the uncertainty transformation, and in ca. 25\% of the cases the LLM classification changes.
The change in classification is significant (McNemars test. GPT-4o: $\chi^{2}$ (1, N= 1086) = 67.63*; LLaMA3: $\chi^{2}$(1, N = 816) = 46.12*; DeepSeek-v2: $\chi^{2}$(1, N = 660) = 50.74*).\footnote{\label{note1}* = p<0.001.}

\begin{table}
\centering
  \begin{tabular}{lllll}
    \hline
    \textbf{Model}           & \textbf{f / nf}   & \textbf{Total}    & \textbf{Proportion f}\\
    \hline
    GPT-4o      &   815 / 271& 1086& .751          \\
    LLaMA3 & 622 / 194 &816& .762\\                          
    DeepSeek-v2 & 477 / 183& 660& .723\\                    
    \hline
  \end{tabular}
  \caption{\label{first_experiment_confusion matrix}
    Confusion matrix depicting classification of transformed statements by different models.
  }
\end{table}

\begin{table}
\centering

  \begin{tabular}{lllllll}
    \hline
               & \textbf{GPT-4o} &  \textbf{LLAMA3}  &  \textbf{Deep}     \\
               & & & \textbf{SeekV2} \\
    \hline
    
          &   f / nf & f / nf & f / nf        \\
    \hline      
    \textbf{Epistemic} & & &   \\                          
    Possible & 141 / 40 & 110 / 26 & 85 / 25 \\
    Probable & 132 / 49 & 102 / 34 & 74 / 36 \\
    \hline
    \textbf{Total} & 273 / 89 & 212 / 60 & 159 / 61 \\
    \hline
    \textbf{Hypothetical} & & &   \\
    Conditional & 145 / 36& 113 / 23& 82 / 23\\
    Doxastic & 101 / 80& 71 / 55& 55 / 55\\
    Question & 152 / 29& 115 / 21& 92 / 18\\
    Weasel & 144 / 37& 111 / 25& 84 / 26\\
    \hline
    \textbf{Total} & 542/182& 410/124& 313/122\\
    \hline
  \end{tabular}
  \caption{\label{confusion_matrix_modalities}
    Confusion matrix depicting classification of transformed statements by different models categorised by modality and linguistic. Note the totals are different per model due to different modelperformance on the control task.
  }
\end{table}

\begin{table}[]
\centering
\begin{tabular}{lllll}
\hline
           & \textbf{Ps. R$^{2}$} & \textbf{Odds}  & \textbf{95\% CI}  \\
           \hline
GPT-4o & 0.195    & 1.071*      & (0.060, 0.078) \\
LLaMA3     & 0.027   & 1.025*      & (0.017, 0.032) \\
DeepSeek-v2 & 0.037   & 1.049*      & (0.035, 0.062) \\
\hline

\end{tabular}
  \caption{\label{logistic_regression analysis}
    Logistic regression statistics *=p < 0.001 
  }
\end{table}

Table~\ref{confusion_matrix_modalities} reports the results of our investigation on the relationship between the change of classification and predictors to which we can expect humans to be sensitive, namely, the 
truth modalities (hypothetical vs epistemic) and 
the linguistic sub-categories in our uncertainty typology.
No significant difference was found between the classification distribution between the truth modalities hypothetical and epistemological ($\chi^{2}$ test of independence. GPT-4o: $\chi^{2}$ (1, N= 1086) = 0.015, \textit{p} = 0.901; LLaMA3: $\chi^{2}$ (1, N = 816) = 0.08, \textit{p} = 0.78; DeepSeek-v2: $\chi^{2}$ (1, N = 660) = 1.0, \textit{p} = 1.0 ). 

For the linguistic subcategories, the $\chi^{2}$ test shows that there is at least one group that is significantly different from the others (GPT-4o: $\chi^{2}$ (5, N=1086) = 49.18*; LLaMA3: $\chi^{2}$ (5, N=816) = 40.88*, DeepSeek-v2: $\chi^{2}$ (5, N=660) = 39.56*).\footnote{* = p<0.001} 
After a pairwise comparison with Bonferroni correction, we found that most linguistic categories cause a change of classification label in around 25\% of the cases with no significant difference (accuracies range between .729--.840 (GPT-4o), .750--.846 (LLaMA3) .673--.836 (DeepSeek-v2)), however, the doxastic category ("I believe", "it is thought" etc.) is significantly causing a non-false classification in around 50\% of the transformed statements independent of model tested (GPT-4o: .52*, LLaMA3: .50*, DeepSeek-v2: .56*).\footnote{* = p<0.001}
For more detailed statistics see \ref{app:additional_statistics_modal_ling}.


Table~\ref{logistic_regression analysis}
presents our logistic regression analysis of the relationship between the elicited frequency and change in factchecking classification. 
We observe that elicited frequency is a significant predictor for classification label in each LLM.






\section{Conclusions and outlook}

In this paper, we explored large language models' 
response to misinformation expressed with uncertainty.
Across LLMs, in about 25\% of the cases a proposition originally classified as false is not longer classified as false when transformed into a statement that expresses uncertainty. Note that in this paper, we do not assume a correct answer to the false/not-false classification of uncertainty-transformed proposition. The reason is that linguistically the transformed sentences have a truth value which is unclear or undefined. Rather, our goal is simply to measure the impact of the certainty transformation on the response of the LLM.
We found no evidence that this difference is explained by either modality or linguistic subcategory.
During an exploratory analysis on frequency, we found a significant correlation between the elicited frequency of a statement and the odds ratio that an LLM classifies a statement as not false. 
Our findings contribute to the broader question of understanding truth attribution by LLMs in the case of uncertainty.


This paper opens an interesting outlook for future research related to the connection between fact and frequency.
First, we find that elicited frequency holds potential for supporting the explanation of factchecking decisions by LLMs. 
Future work should investigate the nature of this relationship in more detail and assess its robustness.
Second, the connection between fact and frequency is reminiscent of the connection between truth assessment and human cognitive heuristics, especially, illusory truth effects. 
Carefully avoiding undue anthropomorphization of LLMs, future work can investigate whether the connection between factchecking and frequency can shed light on how LLMs represent uncertainty internally.


\section{Limitations}
We consider this study to be exploratory and point to three key limitations that must be address by future work.
First, we have studied only verified-false propositions that have also been classified by the LLMs that we studied as false. 
For greater understanding of LLM responses to  information that is expressed with uncertainty, it is important to consider the full range of cases (i.e., include verified-true propositions and also propositions that LLMs have not classified as false.)

Second, we have only studied propositions from the ClimateFever dataset. 
Limitation to this data set helps us to control for topic, but also means that we do not claim broad generalizability.
Further, the ClimateFever data set is about five years old and, because it is publicly available, we are not certain that it is not included in the training data of our large language models.
Note that this point applies to the original verified-false propositions.
The uncertain statements that we tested on the basis of these propositions were created explicitly for the purpose of this paper.

Third, for the classification of misinformation propositions, we have prompted the LLM to classify the uncertain statement as false or true.
In the majority of cases (75\%) the LLM returned false.
For the remaining cases (25\%) it returned true or something else, including a repeat of the question or another statement.
We mapped all output other than false to not-false in this paper.
Future work should further investigate not-false responses, specifically, in cases where uncertainty changes the proposition to be assessed. A concrete example being doxastic cases where the original proposition is encapsulated in another proposition; e.g, "Some people believe the earth is flat" is now a true statement, because it's now about people's beliefs, not whether the earth is flat.
A brief qualitative review of our output did not indicate any interesting trends within this study, but future research could explore this space more elaborately. 

We would like to point out that our work focuses on modality and linguistic subcategory, whereas other aspects of the expression of uncertainty are also interesting.
Specifically, we mention the strength of uncertainty studied by~\citet{zhou_navigating_2023} in a study of answers generated by LLM.
The uncertainty transformations of different strengths our data set, we found, did not support interesting insight on strength. 
We focus here on uncertainty (weakeners), but future work should look at certainty (strengtheners) as well. 
The framework by Rubin et al. \cite{rubin2006certainty} could be helpful in exploring more dimensions of uncertainty. Lastly, we explored the direction of argumentation type as constructed by (\cite{coan2021computer, lamb2020discourses} but found our data not suited for this avenue due to large class imbalances. Future work could explore persuasion and argumentation strategy as a possible explanation for LLMs classification of uncertainty. 

From the technical perspective, we would like to note two limitations.
First, the uncertainty transformations were applied using LLaMa2.
Although they were checked by hand, it could be that a human would have applied them somewhat differently.
Second, the false/not-false prompt was based on a careful study of the literature. 
However, since we are to our knowledge the first to elicit frequency, we formulated the frequency elicitation prompt ourselves.
We chose not to optimize this prompt, since it is not possible to know the correct frequency of the answers and it would be obviously wrong to optimize the prompt to increase the correlation. 
Future work should, however, develop more refined methods for eliciting frequency from large language models.

We conclude with a comment on the broader context of our work.
When an uncertainty transform is applied to proposition, the resulting statement has a truth value that is unclear or undefined.
In this work, we do not take a position about what the correct response of a LLM should be to a false statement expressed with uncertainty. 
The question of how LLMs should handle this case needs to be answered by researchers in the area of psychology and philosophy, and ultimately be encoded in best practices or legally regulated.
We hope that researchers in these fields find the results presented here to be a useful snapshot of how LLM respond to misinformation expressed with uncertainty.

\section{Ethical considerations}
This work uses an openly available data set~\cite{diggelmann_climate-fever_2021} and does not involve a participant study.
Nonetheless, its ethical considerations deserve comment.
Applying uncertainty transformations is an approach that bad actors can exploit to attempt to circumvent flagging by fact checkers who are using LLMs.
Our results show that in the majority of the cases, uncertainty transformations do not change the response of the LLM.
However, it is important to recognize that people should be able to discuss false statements without being flagged by factcheckers.
Uncertainty transformations are used in such discussions.
Assuming across the board that misinformation transformed by uncertainty is still misinformation, could impose undesirable constraints on free speech.

\section{Acknowledgements}
This work was partially supported by the European Union under the Horizon Europe project AI-CODE (GA No. 101135437).
\bibliography{custom.bib}
\newpage
\appendix

\section{Appendix}
\begin{figure}[h]
    \includegraphics[width=0.9\linewidth]{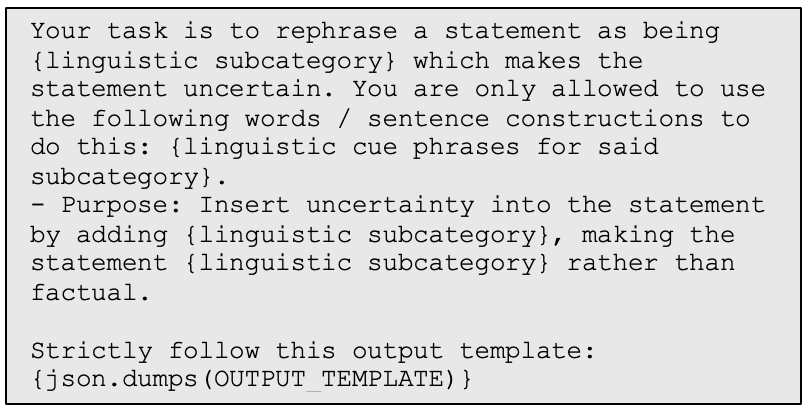}
    \caption{The prompt used for rewriting the propositions into uncertain statements}
    \label{fig:prompt_rewrite}
\end{figure}

\subsection{Transformation}
\label{app-transformation}
We transformed the propositions to uncertain statements using a local version of LLaMA2 to prevent data pollution. We prompted it to, for every linguistic subcategory, rewrite the statements given a set of cue phrases \ref{fig:prompt_rewrite}. We set the temperature to 0.5 to include some creativity, but at the same time contain the model to only use the given cue phrases. The cue phrases were provided by our typology. We manually checked each rewrite to make sure it was 1) felicitous, 2) limited to one linguistic category, 3) conveyed the same information as the original proposition. As an extra control measure, an independent coder used the typology and classified a sample of 250 rewritten statements to specific categories. He correctly appointed each rewritten statement to the correct linguistic subcategory. 
\newpage

\label{sec:appendix}

\subsection{Classification of original propositions}
\label{app-control-class}

\begin{table}[h]
\centering
  \begin{tabular}{lllll}
    \hline
    \textbf{Model}           & \textbf{f / nf}   & \textbf{Total}    & \textbf{Accuracy}\\
    \hline
    GPT-4o      &   181 / 29 & 211 & .862          \\
    LLaMA3 & 136 / 75 &211& .645\\                          
    DeepSeek-v2 & 110 / 101& 211& .521\\                    
    \hline
  \end{tabular}
  \caption{\label{first_experiment_confusion matrix}
    Confusion matrix depicting classification of original propositions statements by different models.
  }
\end{table}

\newpage
\subsection{Additional statistics for modalities \& linguistic subcategories }
\label{app:additional_statistics_modal_ling}
\begin{table*}[h]
    \centering
    \begin{tabular}{llc}
        \hline
        \textbf{Subcategory 1} & \textbf{Subcategory 2} & \textbf{Corrected p-value} \\
        \hline
        Conditional & Doxastic  & 0.000065 \\
        Conditional & Possible  & 1.000000 \\
        Conditional & Probable  & 1.000000 \\
        Conditional & Question  & 1.000000 \\
        Conditional & Weasel    & 1.000000 \\
        Doxastic    & Possible  & 0.000479 \\
        Doxastic    & Probable  & 0.033841 \\
        Doxastic    & Question  & 0.000015 \\
        Doxastic    & Weasel    & 0.000253 \\
        Possible    & Probable  & 1.000000 \\
        Possible    & Question  & 1.000000 \\
        Possible    & Weasel    & 1.000000 \\
        Probable    & Question  & 1.000000 \\
        Probable    & Weasel    & 1.000000 \\
        Question    & Weasel    & 1.000000 \\
        \hline
    \end{tabular}
    \caption{Pairwise comparisons of linguistic subcategories with corrected p-values for GPT4o}
    \label{tab:linguistic_pvalues}
\end{table*}

\begin{table}[h]
    \centering
    \begin{tabular}{llc}
        \hline
        \textbf{Subcategory 1} & \textbf{Subcategory 2} & \textbf{p-value} \\
        \hline
        Conditional & Doxastic  & 0.000519 \\
        Conditional & Possible  & 1.000000 \\
        Conditional & Probable  & 1.000000 \\
        Conditional & Question  & 1.000000 \\
        Conditional & Weasel    & 1.000000 \\
        Doxastic    & Possible  & 0.000722 \\
        Doxastic    & Probable  & 0.206004 \\
        Doxastic    & Question  & 0.000004 \\
        Doxastic    & Weasel    & 0.001362 \\
        Possible    & Probable  & 1.000000 \\
        Possible    & Question  & 1.000000 \\
        Possible    & Weasel    & 1.000000 \\
        Probable    & Question  & 0.116090 \\
        Probable    & Weasel    & 1.000000 \\
        Question    & Weasel    & 1.000000 \\
        \hline
    \end{tabular}
    \caption{Pairwise comparisons of linguistic subcategorys with corrected p-values for LLAMA3}
    \label{tab:treatment_pvalues_2}
\end{table}

\begin{table}[h!]
    \centering
    \begin{tabular}{llc}
        \hline
        \textbf{Subcategory 1} & \textbf{Subcategory 2} & \textbf{p-value} \\
        \hline
        Probable    & Possible    & 0.000519 \\
        Probable    & Doxastic    & 1.000000 \\
        Probable    & Conditional & 1.000000 \\
        Probable    & Question    & 1.000000 \\
        Probable    & Weasel      & 1.000000 \\
        Possible    & Doxastic    & 0.000722 \\
        Possible    & Conditional & 0.206004 \\
        Possible    & Question    & 0.000004 \\
        Possible    & Weasel      & 0.001362 \\
        Doxastic    & Conditional & 1.000000 \\
        Doxastic    & Question    & 1.000000 \\
        Doxastic    & Weasel      & 1.000000 \\
        Conditional & Question    & 0.116090 \\
        Conditional & Weasel      & 1.000000 \\
        Question    & Weasel      & 1.000000 \\
        \hline
    \end{tabular}
    \caption{Pairwise comparisons of linguistic subcategories with corrected p-values for DeepSeek-v2}
    \label{tab:DeepSeek-v2_mod_lin}
\end{table}

\end{document}